\pdfoutput=1

\documentclass[11pt]{article}

\usepackage[]{ACL2023}

\usepackage{times}
\usepackage{latexsym}

\usepackage[T1]{fontenc}

\usepackage[utf8]{inputenc}

\usepackage{microtype}

\usepackage{inconsolata}

\usepackage[inline,shortlabels]{enumitem}

\usepackage{xcolor} 

\usepackage{tcolorbox}

\newcommand{\etc}{etc.\ }
\newcommand{\eg}{e.g., }
\newcommand{\cf}{cf.\ }
\newcommand{\ie}{i.e., }

\newcommand{\figref}[1]{Fig.~\ref{#1}}    
\newcommand{\Figref}[1]{Figure~\ref{#1}}  
\newcommand{\tabref}[1]{Table~\ref{#1}}
\newcommand{\Tabref}[1]{Table~\ref{#1}}
\newcommand{\secref}[1]{Section~\ref{#1}}

\newcommand{\appref}[1]{Appendix~\ref{#1}} 

\usepackage{color}

\newcommand{\thms}[1]{\textcolor[rgb]{.8,0.3,0.1}{#1}}
\renewcommand{\thms}[1]{\textcolor[rgb]{0,0.0,0.0}{#1}}
\newcommand{\semere}[1]{\textcolor[rgb]{1,0.2,0.4}{#1}}
\renewcommand{\semere}[1]{\textcolor[rgb]{0,0.0,0.0}{#1}}
\newcommand{\johannes}[1]{\textcolor[rgb]{0.5,0.5,0.7}{#1}}
\renewcommand{\johannes}[1]{\textcolor[rgb]{0,0.0,0.0}{#1}}

\newcommand{\chris}[1]{\textcolor{violet}{#1}}
\renewcommand{\chris}[1]{\textcolor[rgb]{0,0.0,0.0}{#1}}

\newcommand{\semerenew}[1]{\textcolor[rgb]{1,0.2,0.55}{#1}}
\renewcommand{\semerenew}[1]{\textcolor[rgb]{0,0.0,0.0}{#1}}

\newcommand{\semerecamera}[1]{\textcolor[rgb]{1,0.2,0.55}{#1}}
\renewcommand{\semerecamera}[1]{\textcolor[rgb]{0,0.0,0.0}{#1}}

\usepackage{booktabs}

\usepackage{amsmath}
\usepackage{amssymb}
\usepackage[inline,shortlabels]{enumitem}

\usepackage{multirow}
\usepackage{enumitem}
\usepackage{amsmath} 

\usepackage[colorinlistoftodos,prependcaption,textsize=normalsize]{todonotes}

\usepackage{todonotes}

\usepackage{graphicx}
\graphicspath{ {./figures/} }

\newcommand\roberta{{\texttt{XLM-RoBERTa}}}

\newcommand\FFNN{{\texttt{FFNN}}}

\newcommand{\h}{\mathbf{h}}

\title{Learning from Partially Annotated Data: \\Example-aware Creation of Gap-filling Exercises\\ for Language Learning}

\author{Semere Kiros Bitew$^{*}$, Johannes Deleu$^{*}$, A. Seza Doğruöz, Chris Develder \\ 
and \textbf{Thomas Demeester} \\
IDLab, Ghent University - imec \\
$^{*}$ \small Equal contribution \\
\{\texttt{semerekiros.bitew, as.dogruoz, firstname.lastname\}@ugent.be}\\
}
\begin{document}

\maketitle
\begin{abstract}
\chris{Since performing exercises (including, \eg practice tests) forms a crucial component of learning, and creating such exercises requires non-trivial effort from the teacher, there is a great value in automatic exercise generation in digital tools in education.
In this paper, we particularly focus on automatic creation of gap-filling exercises for language learning, specifically grammar exercises.
Since providing any annotation in this domain requires human expert effort, we aim to avoid it entirely and explore the task of converting existing texts into new gap-filling exercises, purely based on an example exercise, \emph{without explicit instruction or detailed annotation} of the intended grammar topics. 
We contribute 
\begin{enumerate*}[(i)]
    \item a novel neural network architecture specifically designed for aforementioned gap-filling exercise generation task, and 
    \item a real-world benchmark dataset for French grammar.
\end{enumerate*} 
We show that our model for this French grammar gap-filling exercise generation outperforms a competitive baseline classifier by 8\% in F1 percentage points, achieving an average F1 score of 82\%.
Our model implementation and the dataset are made publicly available\footnote{\url{https://github.com/semerekiros/GF2/}} to foster future research, thus offering a standardized evaluation and baseline solution of the proposed partially annotated data prediction task in grammar exercise creation.
}
\end{abstract}

\section{Introduction}
\label{sec:intro}
\chris{While digital education tools have been increasingly developed and deployed for over a decade, the e-learning sector has definitely boomed in the wake of COVID-19, even leading to a new Digital Education Action Plan from the European Commission.\footnote{\url{https://education.ec.europa.eu/focus-topics/digital-education/action-plan}}
As one application in e-learning, we particularly focus on language education, and specifically on the automatic generation of gap-filling grammar exercises.
This type of exercises has been shown to be very effective in language learning, with a noticeable effect of such practice tests on students progress and is generally considered as a global measure of language proficiency \cite{oller1973cloze}.
\semerecamera{Furthermore, automatic generation of exercises has been shown to produce relatively high quality exercises, for example, for multiple choice questions~\cite{mitkov2006computer},}
demonstrating the potential effectiveness of reducing human effort and offering cost-effective solutions towards personalized exercise generation.
In terms of technology, recent developments in natural language processing, \eg BERT~\cite{devlin2018bert}, GPT-3~\cite{brown2020language}, InstructGPT~\cite{ouyang2022training}, open up new opportunities for further upscaling and improving automatic generation of such tests/exercises.
}

\chris{In this paper we specifically propose to generate grammar exercises from existing texts, by inducing well-chosen gaps in a given input sentence, following a set of given example exercise sentences.
Further, we aim to create models that can be trained on the exercises themselves, without further annotations. 
The latter implies that we want to forgo a \semerecamera{fully} supervised learning setting, because such models would require \semerecamera{each gap in} the available exercises to be manually annotated with additional metadata, such as the particular exercise type, \eg for gap-filling exercises, a suitable category such as a verb tense. 
Thus, we focus on converting given input texts into gap-filling exercises, by mimicking the implicit rules underlying a given example exercise, rather than by following explicit instructions such as a prescribed exercise type.
}

\paragraph{Application scenario:} 
Consider a language teacher, who just introduced a particular grammatical topic (e.g., a new verb tense), and needs the students to practice.  The grammar topic of interest may need to be practiced in combination with particular other topics (e.g., related tenses already studied by the students).  Given that gap-filling questions can be completed online and automatically assessed~\cite{daradoumis2019analyzing}, the teacher creates a new gap-filling exercise, covering these combined grammar topics. 
The goal of our model is then 
to support the automatic creation of new exercises, based on that example exercise, by transforming other texts provided by the teacher into additional gap-filling exercises that target the same linguistic topics to be practiced, without explicit instructions by the teacher of which topics the model should include.
This would allow the teacher to rapidly create new training material for the students, potentially more diverse, for example, in terms of topics of the texts, their temporal relevance, or the inherent linguistic difficulty.

\paragraph{Learning from partially annotated data:}
The scenario outlined above represents a learning task in between one-shot learning (\ie learning from one example~\cite{wang2020generalizing} and full supervision (\ie based on the full annotation of all examples).
On the one hand, the one-shot setting considers the example exercise as a single training instance defining the nature of the prediction task by the way it was constructed by the teacher (in this case, the included grammar topics). \chris{On the other hand, the fully supervised setting}
would require at least explicit knowledge of all exercise instructions (\ie gap types per exercise).
\chris{Although we assume the availability of an entire corpus of such exercises, on overlapping grammar topics, we will not rely on explicit annotation of the nature of the gaps (\ie gap type that defines the type/scope of the grammar exercise, or even just identifying the word category).
Thus, we do want to learn from partially annotated examples, where the annotation is limited to just the indication of the gap and the text span that constitutes the expected answer. 
This basically amounts to the type of information that would be available in a one-/few-shot setting, but we aim to leverage the complete corpus to train our models.}

\chris{Note that, while creating exercises, teachers are aware of the envisioned exercise type and the gap types, and such exercise type would also be communicated (\eg as a free-text instruction) to students.
Still,}
to keep our experiments and the gained insights transparent, we left out any exercise level instructions for our experiments.

\begin{figure*}[!tbh]
\includegraphics[width=\textwidth]{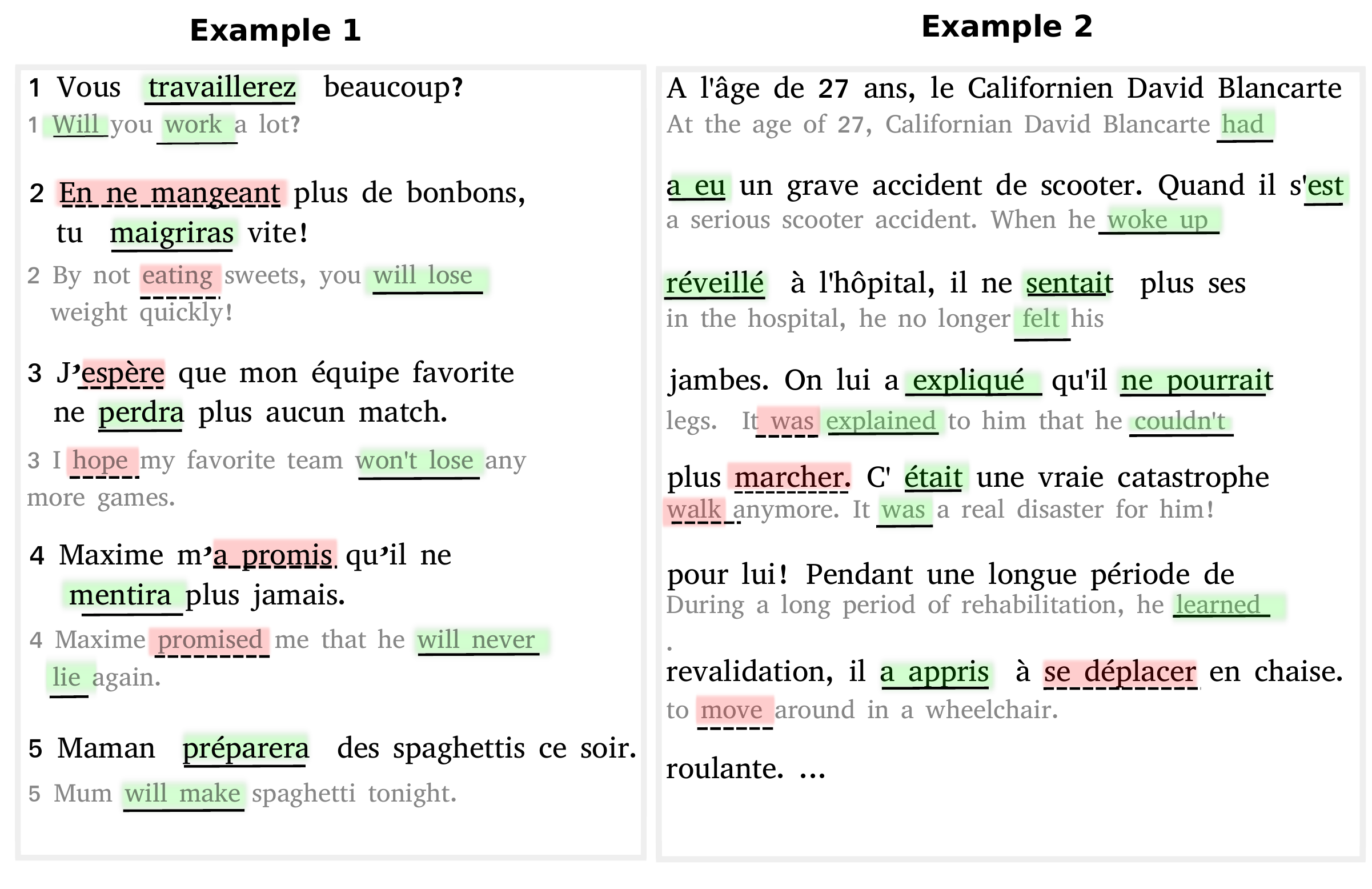}
\caption{\thms{French grammar exercise from the GF2 corpus, with English translations for convenience shown in light grey. Green spans (with solid underline) are actual gaps as selected by teachers in the dataset, red spans represent potential gaps on other grammar topics but were not marked as gaps. (Left) Isolated sentence exercise  with focus on a single tense (\emph{futur simple}); (right) full text exercise combining two tense types (\emph{imparfait} and \emph{pass\'e compos\'e}). 
}
} \label{fig:data_examples}
\end{figure*}

\paragraph{Link with related research:}

In broad terms, the proposed work fits within the area of automatic question generation (AQG) for the educational domain. In the field of education, creating questions manually is an arduous task that demands considerable time, training, experience, and resources from educators~\cite{davis2009tools}. As a solution to this challenge, researchers have turned towards AGQ approaches to automatically generate homework, test, and exam exercises from readily available plain text that requires little to no human calibration. In particular, educational AQG systems have been developed for generating \emph{factoid questions} covering several subjects such as history~\cite{al2011ontoque,papasalouros2008automatic}, general sciences~\cite{sun2018automatic,stasaski2017multiple,conejo2016siette}, health and biomedical sciences~\cite{pugh2016using,afzal2014automatic}, etc., as well as for \emph{language learning} such as vocabulary or grammar exercises~\cite{susanti2017evaluation,hill2016automatic,goto2010automatic}. There has been some more generic recent work, however, on finding distractors for multiple choice questions across subjects and languages \cite{bitew2022learning}. It is line with recent work on training deep neural networks for general-purpose question generation~\cite{du-etal-2017-learning}, based on large training sets. 
There is a clear preference for two question types that allow for automated assessment, i.e., multiple-choice questions (e.g., in \cite{stasaski2017multiple,pugh2016using,afzal2014automatic,papasalouros2008automatic}) or gap-filling questions (as in \cite{hill2016automatic,malinova2016automatic,perez2012generating,goto2010automatic}). 

\semerenew{Our work is focused on gap-filling questions, which typically require test-takers to fill in blank spaces in a text with missing word(s) omitted by test developers. The missing words can either be chosen from a set of possible answers (\ie closed cloze questions), or generated from scratch using hints provided in the text (\ie open cloze questions). To generate such questions, various strategies were employed, such as deleting every nth word from a text~\cite{taylor1953cloze}, or rationally deleting words according to specific purpose, \eg usage of prepositions~\cite{lee2007automatic}, verbs~\cite{sumita2005measuring} \etc Previous studies have relied on selecting informative sentences~\cite{slavuj2021automatic,pino2008selection} from existing corpora, such as textbooks~\cite{agarwal2011automatic}, WordNet~\cite{pino2008selection}, and then using techniques such as POS tagging~\cite{agarwal2011automatic} or term frequency analysis~\cite{mitkov2006computer} to determine gap positions. More recently, \citet{marrese2018learning}, have developed sequence labeling model to automate the process of generating gap-filling exercises.}

\semerecamera{Another very relevant work by \citet{felice-etal-2022-constructing} devised a method to adapt an ELECTRA~\cite{Clark2020ELECTRA:} model for the purpose of generating open cloze grammar exercises in English. Their approach involved classifying each individual token as either a gap or non-gap. However, there exist several notable distinctions between their approach and our own. Firstly, unlike their method that solely focused on individual tokens, we make gap decisions based on spans. This distinction is essential as our gaps can encompass multiple words, allowing for more comprehensive and contextually accurate grammar exercises. Secondly, our objective and experimental setup differ significantly. Our ultimate goal is to generate multiple versions of the same text, with each version targeting a distinct grammar aspect (\eg future tense, prepositions of time or combinations of different types). In contrast, their approach consistently produces exercises of the same type for a given input text (\ie similar to our baseline model), lacking the versatility and adaptability our model offers.}

We observed a tendency in generation of gap-filling questions aiming at well-defined tasks. To the best of our knowledge, none of the prior works have proposed strategies to capture common underlying structures in terms of task definition, while training on a heterogeneous set of real-world examples (e.g., covering various grammatical topics).

\paragraph{Key research contributions: }

\begin{itemize}
    \item We introduce the task of the example-aware prediction of suitable linguistic gaps in texts \semerecamera{based on partially annotated data. This task is of paramount importance in the development of new gap-filling exercises.}
    \item We present our real-world dataset of French gap-filling exercises covering unknown combinations of grammatical aspects. Our dataset called GF2 (\textit{`Gap-Filling for Grammar in French'}) is released as a research benchmark for the introduced task.
    \item We propose and train a suitable neural network architecture for the task, and show that conditioning the model's output for a given input text on an example exercise of the envisioned exercise type, leads to an increased effectiveness, compared to an example-independent baseline model. 
    Additionally we analyse the model's ability to disentangle elementary exercise types, without being explicitly trained to do so, and \semerenew{we observe that it can recognize types to some extent, especially for the most commonly occurring types in the test set.}

\end{itemize}

\section{Gap-filling Exercise Creation as a Span Detection Task}
\label{sec:task}
\thms{This section describes the particular prediction task this paper focuses on. We cast the creation of a French gap-filling exercise from an input text as a \emph{binary span detection task}: the goal is detecting each span (i.e., consecutive sequence of tokens) that represents a correct gap. For clarity, we left out creating the `hint' (e.g., the infinitive for verbs) which would make it a finalized gap-filling exercise, as it is considered less challenging and may deviate attention from the core problem of identifying the correct spans.}

\thms{
\Figref{fig:data_examples} shows two example gap-filling exercises, with indication of the ground truth spans in green (and with solid underline). We denote the distinguishing feature of each gap as its \emph{gap type} (e.g., the tense \emph{futur simple} for each of the valid tags in Example 1). An exercise typically covers multiple gap types, and the particular combination that characterizes a given exercise is called its \emph{exercise type}. 
\thms{As such, many different exercise types can be constructed, and some may be unseen in the training data.} 
For example, Example 2 (again in \figref{fig:data_examples}) combines three tenses (\emph{imparfait}, \emph{pass\'e compos\'e}, and \emph{conditionnel présent}), which constitutes its exercise type. However, the same text could have been enriched with different gaps, corresponding to a different exercise type. In fact, our test set of one hundred exercises, for which we annotated gap types in terms of 12 elementary verb tenses, covers a total of 35 such composite exercise types. 
} 

\semerecamera{Considering the lack of information regarding the exercise types for the training exercises, we further define the task we are examining more precisely. The objective is to detect the valid spans (\ie spans that will be designated as gaps) of a given flat \emph{input} text that mimics the same underlying exercise type as an example gap-filling exercise, which we denote as the \emph{exemplar}. This exemplar serves as an indirect reference for the model to understand the desired exercise type. By utilizing this approach, we can better inform the model about the desired exercise type while accounting for the the lack of exercise information available. }


\thms{Note that our goal is working with real-world data. Our training data contains gap-filling examples following particular unknown exercise types.  Moreover, teachers appear 
\chris{to not} always select every possible span that satisfies the exercise type. We saw cases in our dataset (\cf\secref{sec:data}), where the same verb occurring twice in the same form would be selected as a valid gap only once. Such real-world `inconsistencies' contribute to the challenging nature of learning from such data without additional annotations.}

\section{Example-aware span detection model}
\label{sec:proposedmodel}

\thms{This section describes our baseline model and proposed example-aware gap detection model.}
%
\begin{figure*}[h!]
\includegraphics[width=\textwidth]{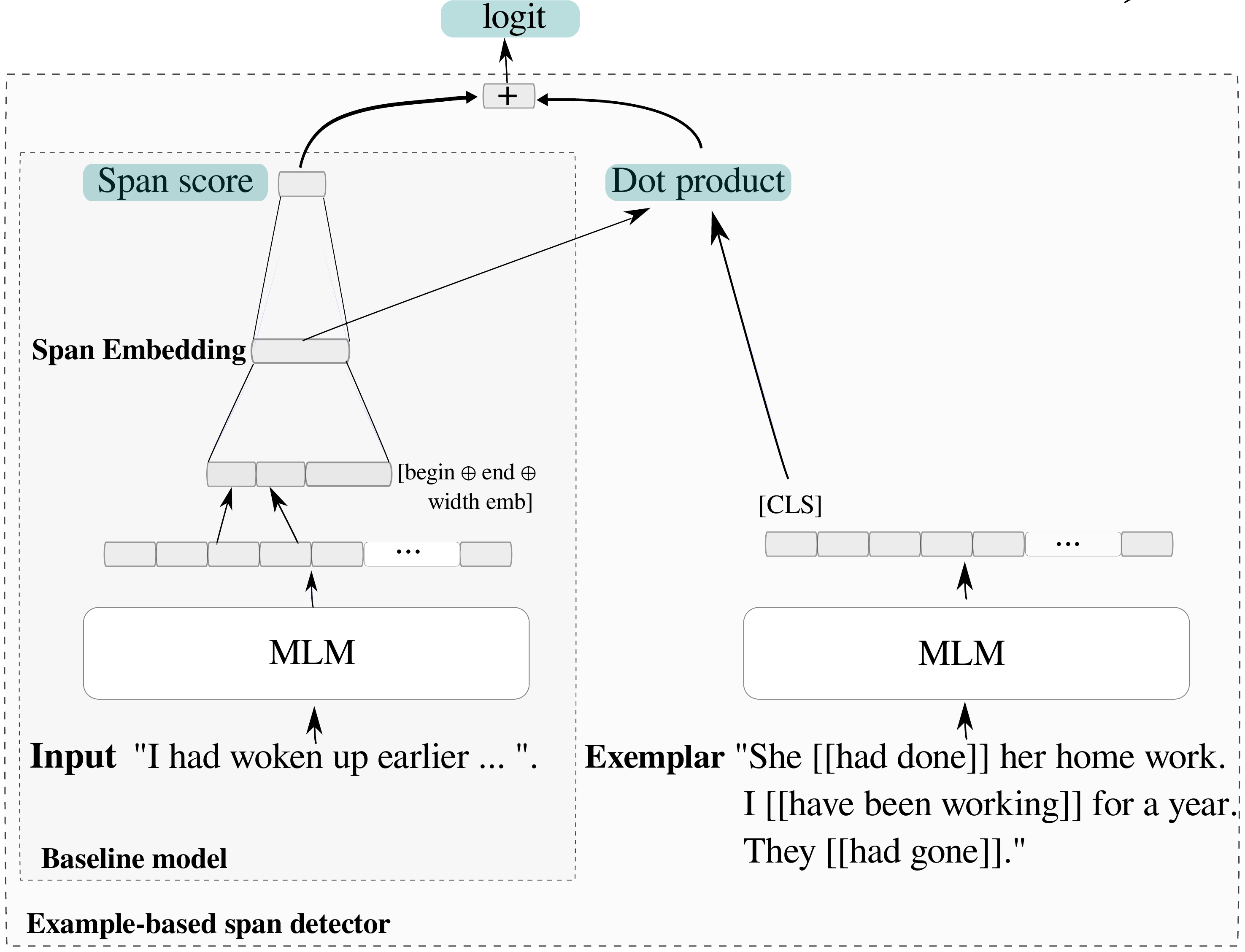}
\caption{Example-aware gap detection model architecture. $\oplus$ denotes concatenation. In general, the model considers all possible spans up to a maximum width, but we depict here only one span from the input for brevity.} \label{fig1}
\end{figure*}
\Figref{fig1} \thms{provides a schematic overview. We first detail the part indicated as \johannes{\emph{Baseline model}}, inside the smaller dashed box, followed by the part that encodes the exemplar, which leads to the full model.
}

\paragraph{Baseline model:}

\thms{An input text $\mathbf{t}$, consisting of $N$ tokens $\mathbf{t}=[t_0, t_1, \ldots,t_{N-1}]$
is encoded by a transformer based masked language model (MLM), in our experiments the multilingual \roberta{}~\cite{conneau2019unsupervised}. From the corresponding transformer outputs $[\h_0, \h_1, \ldots,\h_{N-1}]$ (with $\h_i\in \mathbb{R}^k$, i=0\ldots N-1), vector representations are constructed for all possible spans inside the input sequence, up to a certain length (in our experiments 12 tokens).  The goal is then to make a binary prediction in terms of valid gaps, for each of these spans. In particular, for a span $\varsigma = [t_{\text{start}},\ldots,t_{\text{end}}]$ with \johannes{endpoint tokens $t_{\text{start}}$, $t_{\text{end}}$ }and width $\vert\varsigma\vert = {\small{(\text{end}-\text{start}+1)}}$ in the input text, the corresponding span representation $\h_\varsigma$ is constructed as 
\[\h_\varsigma = \FFNN\big(\h_\text{start} \oplus \h_\text{end} \oplus \h_{\vert\varsigma\vert}\big) \]
in which $\oplus$ represents vector concatenation, $\h_{\vert\varsigma\vert}$ corresponds to a span width embedding, jointly learned with the model, and $\FFNN$ is a fully connected feed-forward model with a single hidden layer, ReLU activation, and output dimension $k$. 
The \roberta{} output representations $\h_\text{start}$ and $\h_\text{end}$ of the start and end token of $\varsigma$ are concatenated with the span width embedding $\h_{\vert\varsigma\vert}$, and transformed through $\FFNN$ into the $k$-dimensional span representation $\h_\varsigma$. 
The probability of span $\varsigma$ representing a valid gap is modeled as
\[p_\text{base}(\varsigma) = \sigma\big(\mathbf{w}\cdot \h_\varsigma +b\big)\]
in which the trainable parameters $\mathbf{w}$ and $b$ are a $k$-length coefficient vector and bias, respectively, $\sigma$ is the sigmoid function, and $\cdot$ represents the dot product.
The baseline model is trained by minimizing the cross entropy loss between each span's score $p_\text{base}(\varsigma)$ and its label (1 for valid gaps, 0 otherwise). At inference, spans are predicted as gaps as soon as $p_\varsigma\geq 0.5$.
}

\paragraph{Example-aware gap detection model:} As shown in \thms{\figref{fig1}, our example-aware model is a direct extension of the baseline model which by construction makes example-unaware predictions.
The same MLM that encodes the input, is now used to also encode the exemplar, which contains the example exercise text as well as the correct gap information.  The latter is added by surrounding each gap with the special tokens `[[' and `]]' (as seen in the figure).
Details on how the examples are chosen, are provided in \secref{sec:trainingandinference}. 
The exemplar representation $\h_{exemplar}$ is obtained as the MLM's [CLS] representation\footnote{[CLS] is a special token that is prepended to the input, and its corresponding output representation is pretrained to represent the entire sequence that is used for classification tasks}. 
}

\thms{We then quantify the compatibility of each span $\varsigma$ in the input text with the exemplar, through the dot product $\h_{\text{exemplar}} \cdot \h_\varsigma$ of their respective representations. In a direct extension of the baseline model, it leads to the proposed model for the probability $p_\text{example-aware}(\varsigma)$ that $\varsigma$ represents a valid gap:
\[p_\text{example-aware}(\varsigma) = \sigma\big(\h_\varsigma\cdot \mathbf{w} + \h_\varsigma\cdot \h_{\text{exemplar}} + b\big)\]
}

\section{Empirical validation on real-world data}
In this section, we first introduce the dataset that we will publicly release. Then, we explain how we train our models and use them for inference. Finally, we describe the strategies we adopted to evaluate the effectiveness of our models.

\subsection{GF2 dataset: \bf{G}ap-\bf{F}ill for \bf{G}rammar in \bf{F}rench}
\label{sec:data}



\thms{We denote our new dataset as ``Gap-Filling for Grammar in French'' (GF2). It was contributed by Televic Education\footnote{\url{https://www.televic.com/en/education}}, and gathered through its education platform 
assessmentQ\footnote{\url{https://www.televic-education.com/en/assessmentq}}.
}
AssessmentQ is a comprehensive online platform for interactive workforce learning and high-stakes exams. It allows teachers to compose their questions and answers for practice and assessment. As a result, the dataset is made up of a real-world set of gap-filling grammar exercise questions for French, manually created by experts. 
We cleaned and preprocessed the data before we could use it to train our models. \thms{First, organizational metadata information was removed. Other elements that we removed are the hints within the body of the text that could easily give away the gap positions, as well as inline instructions (if present) about the exercise type.} Second, we automatically stripped off HTML tags from the documents. 
Our final dataset contains a total of 768 exercise documents, in which a total of 5,530 spans are tagged as gaps. 
The exercises were randomly split into 618 train documents, and 50 and 100 for validation and test, respectively. \Tabref{tab:statistics} summarizes FG2's descriptive statistics. 

For the validation and test exercises, we made an extra manual effort to enrich each of the existing gaps with their gap type. Our annotations reflect the fact that the data contains a mix of verb and non-verb gaps. Every gap has an associated word type attribute (e.g. adverb, adjective, verb) and in case of verbs a tense attribute. In what follows we zoom in on the verb gaps and consider the tense as the main gap type. The bottom half of \tabref{tab:statistics} shows the frequency of occurrence for the main verb types in the development and test documents. 
We use this annotations to get insights into the dataset and to evaluate the properties of our models (see \secref{sec:resultsanddiscussion}). 
Note that the examples shown in \figref{fig:data_examples} are actual entries from the GF2 dataset.

\begin{table*}[h]
\centering
\caption{Statistics of the FG2 dataset and breakdown into key verb tenses (gap types) in the validation and test split. For the train split we only know gap spans, not their types, since they are not labelled.}\label{tab:statistics}
\begin{tabular}{lrrr}
\toprule
 &  \bf{Train} & \bf{Dev} & \bf{Test}\\
\midrule
\# Documents  &  618 & 50 & 100 \\
\# Sentences & 4786 & 378 & 707\\
\# Gaps &  4518 & 365 & 647 \\
\midrule

Subjonctif Pr\'esent (SPR) & \sc{UNK} &   1  & 28\\
Pass\'e Compos\'e (participe pass\'e) (PCP)$\qquad$   & \sc{UNK} &  31   & 8 \\
Pass\'e Composé (PC)                & \sc{UNK} &   84  & 108 \\
Imparfait (IM)                   & \sc{UNK} &    8    & 46 \\
Conditionnel Pr\'esent (CPR)       & \sc{UNK} &   23  & 92 \\
Pass\'e R\'ecent (PR)                 & \sc{UNK} &   0  & 12 \\
Futur Proche (FP)                   & \sc{UNK} &  1   & 9 \\
Futur Simple (FS)                  & \sc{UNK} &   8  & 49 \\
Indicatif Présent (IP)            & \sc{UNK} &  126  & 144 \\
Conditionnel Pass\'e (CPA)          & \sc{UNK} &    0    & 3 \\
Imp\'eratif (IMP)                   & \sc{UNK} &   12   & 26 \\
Plus-que-parfait (PQ)             & \sc{UNK} &   0  & 1 \\

\bottomrule
\end{tabular}
\end{table*}

\subsection{Training and inference}
\label{sec:trainingandinference}


\semerecamera{Our baseline model is relatively straightforward to train. We designate all spans indicated as gaps in our training data as valid gaps, which are considered positive examples. Conversely, any spans that are not indicated as gaps are labeled as negatives. We train our model by minimizing the cross entropy loss between each span's predicted score and its label as described in \secref{sec:proposedmodel}. However, training our example-aware model poses a challenge due to the lack of knowledge regarding the exercise types of the training exercises. Using one exercise as an example and another exercise of the same type as the input, along with the corresponding targets, is not therefore feasible. Instead, we make the assumption that exercises are generated by teachers who consistently follow the underlying exercise type throughout the entire exercise. As a result, we divide the training exercises into two parts: one part is used as an exemplar, and the other part serves as the actual input, for which the gaps are assumed to follow the same exercise type.}


\semerecamera{To this end, we first segment each document in the training set into a list of sentences, along with their corresponding target gap positions. We create a new (exemplar, input) training pair by sampling one sentence to be used as the input, and uniformly sampling one up to $m$ sentences from the remaining sentences within the same document to be used as the exemplar. The exemplar is constructed by concatenating these sampled sentences, with the addition of special symbols denoting the gap locations. (See \appref{sec:appendix} for details.)}
\thms{These are the positive training examples that encourage the model to correctly learn predicting example-aware gaps. However, to facilitate efficient learning, 
it is crucial to also provide negative examples on which the model should not predict gaps. To create such negative training instances, a sentence is sampled as input from the considered document, but its span targets are set to zero (no gaps), and the negative exemplar is composed as before (including indicating the gaps), but by sampling sentences from a randomly selected \emph{other} training exercise. There is risk of incidentally creating false negative training examples, if the exemplar gaps correspond with left-out gaps in the input.  However, negative exemplars appeared important for obtaining a suitable model.
}

We determine the optimal proportion of negative to positive instances for training our models by employing a fine-tuning approach utilizing the macro F1 score as the evaluation metric on the validation set. \thms{This increases the impact of the rarer gap types in the metric, and therefore in the final model, which we considered important for practical use. Other choices could have been made, however.}
Ultimately, the final model is trained on the union of the training and validation splits, using the optimal proportion determined via the fine-tuning process.
  


\semerenew{During inference, we use our trained model to predict the gap positions for an input text that is implicitly conditioned on the target exercise type through the exemplar.} 

\paragraph{Implementation and training details:} We implement our models using pytorch and Huggingface. We initialize our MLM encoders with \texttt{xlm-roberta-base}. To avoid extensive hyper-parameter tuning, we made the following choices; a learning rate of 2e-5 in combination with the robust Adam optimizer. We use a batch size of 16 and train our models for 30 epochs. We consider all spans up to a maximum length 12 and we set \emph{k}, the number of sentences per exemplar to 3.

\begin{table*}[tbh!]
\centering
\caption{\semere{Tense disentangling ability in terms of precision, recall, and F1 (in \%) on the test set, as reported for each key verb tenses (with on the right their support, i.e., number of occurrences). We also show the macro F1 score for the static baseline (\emph{baseline}) and our proposed example-aware gap prediction} (\emph{ours}).}\label{tab:disentangling}

\begin{tabular}{l@{\hskip 0.2in}r@{\hskip 0.2in}r@{\hskip 0.2in}r@{\hskip 0.2in}r@{\hskip 0.2in}r@{\hskip 0.2in}r@{\hskip 0.2in}r}
\toprule
& &\multicolumn{1}{c}{\emph{Baseline}} &  \multicolumn{4}{c}{\emph{Ours}} \\
\cmidrule(lr){2-4} \cmidrule(lr){5-7} 
Tenses &P &R & F1  &P &R &F1  &Support\\
\midrule
SPR             & 5.0\tiny{$\pm$0.3}        & 78.6\tiny{$\pm$8.9}       & 9.4\tiny{$\pm$0.6}        & 7.5\tiny{$\pm$0.2}    & 81.0\tiny{$\pm$12.5}        & 13.7\tiny{$\pm$0.4}   & 28 \\
PCP$\qquad$     & 0.1\tiny{$\pm$0.1}        & 4.2\tiny{$\pm$6.3}        & 0.2\tiny{$\pm$0.3}        & 12.6\tiny{$\pm$4.1}   & 62.5\tiny{$\pm$12.5}      & 20.7\tiny{$\pm$6.2}   & 8 \\
PC              & 21.3\tiny{$\pm$1.2}       & 86.4\tiny{$\pm$3.7}       & 34.2\tiny{$\pm$1.8}       & 64 \tiny{$\pm$9.4}    & 86.1\tiny{$\pm$1.9}       & 73.1\tiny{$\pm$5.5}   & 108 \\
IM              & 9.3\tiny{$\pm$0.4}        & 88.4\tiny{$\pm$3.7}       & 16.2\tiny{$\pm$0.8}       & 12.0\tiny{$\pm$2.5}     & 78.3\tiny{$\pm$10.9}      & 20.9\tiny{$\pm$3.9}   & 46 \\
CPR             & 19.9\tiny{$\pm$0.5}       & 94.5\tiny{$\pm$2.9}       & 32.8\tiny{$\pm$0.8}       & 28.3\tiny{$\pm$2.9}   & 92.4\tiny{$\pm$4.7}       & 43.2\tiny{$\pm$3.1}   & 92 \\
PR              & 2.7\tiny{$\pm$0.1}        & 100.0\tiny{$\pm$0.0}      & 5.3\tiny{$\pm$0.1}        & 9.7\tiny{$\pm$2.0}    & 100.0\tiny{$\pm$0.0}      & 17.7\tiny{$\pm$3.3}   & 12 \\
FP              & 1.6\tiny{$\pm$0.0}        & 77.7\tiny{$\pm$0.0}       & 3.1\tiny{$\pm$0.1}        & 6.0\tiny{$\pm$0.9}      & 77.8\tiny{$\pm$0.0}     & 11.1\tiny{$\pm$1.5}   & 9 \\
FS              & 9.9\tiny{$\pm$0.3}        & 88.5\tiny{$\pm$1.7}       & 17.8\tiny{$\pm$0.5}       & 13.6\tiny{$\pm$1.1}   & 84.4\tiny{$\pm$10}        & 23.3\tiny{$\pm$1.7}   & 49 \\
IP              & 24.6\tiny{$\pm$1.2}       & 75.0\tiny{$\pm$4.3}       & 37.1\tiny{$\pm$1.9}       & 32.0\tiny{$\pm$1.4}     & 66.2\tiny{$\pm$11.9}      & 42.9\tiny{$\pm$2.4}   & 144 \\
CPA             & 0.1\tiny{$\pm$0.1}        & 11.1\tiny{$\pm$16}      & 0.2\tiny{$\pm$0.3}        & 0                     & 0                          & 0                     & 3 \\
IMP             & 5.2\tiny{$\pm$0.3}        & 88.5\tiny{$\pm$2.2}       & 9.9\tiny{$\pm$0.5}        & 16.8\tiny{$\pm$1.7}  & 84.6\tiny{$\pm$3.9}       & 25.3\tiny{$\pm$2.1}  & 26 \\
PQ              & 0.2\tiny{$\pm$0.0}        & 100.0\tiny{$\pm$0.0}      & 0.5\tiny{$\pm$0.0}        & 0.6\tiny{$\pm$0.1}    & 100\tiny{$\pm$0.0}        & 1.2\tiny{$\pm$0.2}    & 1 \\
\midrule
\textbf{Macro F1} &  &\multicolumn{1}{c}{13.9} & &  \multicolumn{4}{c}{\textbf{24.4}} \\
\bottomrule
\end{tabular}
\end{table*}

\subsection{Evaluation setup}
\label{sec:evaluationsetup}
In order to assess and analyze the performance of the baseline and the example-aware model, we design two evaluation strategies that look at different effectiveness aspects.
\paragraph{Binary gap prediction evaluation:} the primary objective of our model is to mimic the real-world setting where gap labels are not given. We measure how well our models predict gap positions (\ie gap or no-gap decisions for all input spans). To do this, we split up each of the exercise documents in our test into two parts that are roughly the same size, \thms{given that by \johannes{assumption} they then represent the same exercise type}. 
We calculate the automated metrics by using one
half as the exemplar and the second as the input text to our model. We repeat this process by exchanging the roles of the parts. It is worth noting that we excluded one-sentence test documents (\semerenew{\ie because they can not be chunked into two parts}), which amount to 16\% of the total test documents. However, since most of the excluded sentences (i.e., one-line documents) only had one gap, we only removed 2.7\% of the total gaps in the test set.

\paragraph{Gap type disentangling evaluation:} The goal of the second evaluation setting is to analyze how well the model has learned to disentangle individual gap types, despite not being explicitly trained to do \johannes{so}. \thms{This analysis is based on the assumption that a model that scores high on that aspect, would be stronger in dealing with \johannes{new or }rare exercise types. \johannes{Potentially even at creating new combinations of existing exercises.} This is an aspect we plan to study further when designing more advanced models in future research.}
To this end, we construct a small set of 12 exemplars, one for each of the key verb tenses, by randomly selecting them from the original data and subsequently removing them from the train/validation/test splits. Each exemplar comprises multiple sentences, all of which are homogeneously annotated with the same intended verb type, which will serve as the desired homogeneous exercise type. 
We evaluate our model on \thms{every sentence of} the test set, \thms{by prompting it with each of these 12 fixed exemplars.}
Based on the gap types we annotated on the test set, we can then compute the precision, recall and F1 score for each of these 12 tenses. 

\section{Experimental Results}
\label{sec:resultsanddiscussion}


In this section, we provide evidence of the effectiveness of our proposed model by reporting and discussing the experimental results. \Tabref{tab:results} summarizes the binary gap prediction evaluation of the baseline vs.~the example-aware model on the test set. We report our results as the mean and standard deviation over five runs, each using a different random seed for model training. The proposed example-aware model (denoted as \emph{ours}) consistently outperforms the example-unaware \emph{baseline} on all metrics. In general, there is an absolute gain of 8 percentage points in F1 for the proposed model in comparison with the baseline, achieving an average F1 score of 82.4\%. This confirms our intention when designing the model, that providing example exercises leads to an increased effectiveness in terms of predicting gap positions compared to the static baseline model.

\begin{table}[!h]
\centering
\caption{Overall binary gap prediction in terms of precision, recall, and F1 (in \%) on the test set. Results shown for the static baseline (\emph{baseline}) and our proposed example-aware gap prediction (\emph{ours}).}\label{tab:results}
\begin{tabular}{llll}
\toprule
 &  \bf{Precision} & \bf{Recall} & \bf{F1}\\
\midrule
\emph{Baseline} &  74.87\tiny{$\pm$2.44} &73.11\tiny{$\pm$2.00} & 73.92\tiny{$\pm$0.49} \\
\emph{Ours} & 84.30\tiny{$\pm$1.70} & 80.74\tiny{$\pm$1.80}  & 82.40\tiny{$\pm$0.20}\\

\bottomrule
\end{tabular}
\end{table}
In \tabref{tab:disentangling}, we show the evaluation of our models in their ability to disentangle the 12 main verb types. We observe that for the tenses with relatively higher support, the example-aware model outperforms the baseline with certainty as demonstrated by the individual F1 scores. 

The overall macro F1 score for the example-aware model stands at 24.4\%, which is low in absolute value, but considerably higher than
the baseline's macro F1 score of 13.9\%. 
We observe that the proposed model is able to recognize verb types such as pass\'e compass\'e (PC), imparfait (IM), and conditionnel pr\'esent (CPR) to some extent with F1 scores of 73\%, 43\%, and 42\%, respectively.  
\thms{However, the low overall scores are not unexpected, because the models are not trained to recognize gap types. Furthermore, some tenses are either very rare (e.g., PQ, CPA, PCP) as \semerenew{indicated by their support}, or may appear mainly in combination with other exercise types. This makes achieving a better resolution in disentangling gap types without any explicit gap labels during training an inherently difficult task.}

\section{Conclusion}
\label{sec:conclusion}

In this paper, we introduced a new task within the general challenge of training models to automatically create new exercises for use in education, based on existing exercises and without requiring additional manual annotations.

In particular, we introduced a dataset and associated prediction task, focusing on detecting gaps within a given input text, without knowledge of the exact exercise type, by only relying on an example exercise. 
We proposed an example-aware neural network model designed for this task, and compared it with a baseline model that does not take into account any example of the desired exercise type. We found that our example-aware model outperforms the baseline model not only in predicting gaps, but also in disentangling gap types despite not being explicitly trained on that task. 
Our real-world GF2 dataset of French gap-filling exercises will be publicly released together with the code to reproduce the presented empirical results.

The presented work fits with our pursuit towards supporting personalized learning experiences by \johannes{either suggesting existing or} generating \johannes{new} exercises that are tailored to students' needs. 
Teachers could also benefit from an increased efficiency in creating new exercises. For example, they could make many and diverse drill and practice exercises on chunks of text based on existing standard exercise types without having to provide extra metadata information such as instructions. We hope our benchmark dataset and task will spark new research in the CL and Educational NLP community.

\section*{Limitations}
\semerenew{We identify two limitations of the current work and make suggestions for future directions. First, while our proposed method is language-agnostic in principle, our evaluation is limited to our French benchmark dataset. Expanding our approach to encompass other languages would bring new and interesting challenges for further investigation. Second, despite topic diversity within our exercise documents (\eg the first example in \figref{fig:data_examples} consists of independent sentences, while the second is a coherent text centered around the same topic.), it would be interesting to quantify the degree of topical bias introduced during our training process and its impact on our binary task evaluation. For future work, we first aim to adapt seq2seq models for our task particularly text-to-text models such as T5~\cite{raffel2020exploring}. There is also potential to explore different prompting strategies for large language models (LLMs), when generating gap-filling grammar exercises. For instance, the utilization of chain-of-thought prompting~\cite{wei2022chain}, which involves generating intermediate steps before producing the final response, could be explored for generating grammar exercises.
Additionally, an interesting future study would involve investigating the number of example demonstrations that LLMs require in order to accurately mimic example gap exercises.}


\section*{Ethics Statement}

\semerenew{In this research, we posit that the dataset and models introduced are of low-risk in terms of potential harm to individuals. The dataset used is a curated selection of existing educational content enriched with meta-data, and we are confident that our compilation of the dataset has not introduced any additional ethical risks. However, it is crucial to emphasize the need for accountability and the establishment of clear guidelines for the deployment of grammar generation models, such as the ones benchmarked in this paper, for educational purposes.}

\semerenew{It should be noted that our models are derived from general-purpose neural language encoders that have been trained on real-world data, which may contain biases or discriminatory content~\cite{bommasani2021opportunities}. As a result, our models may have inherited some of these biases and could potentially base their prediction on such biased information. Therefore, it is imperative for educators and researchers to thoroughly consider these ethical issues and ensure that the generated grammar questions align with educational goals and do not perpetuate harmful biases.}

\semerenew{Educators should retain the final authority in accepting or modifying grammar question suggestions generated by such models, keeping their educational goals in mind (\eg in terms of formative and especially summative assessment). In practice, these models are designed to enhance teachers' efficiency in preparing teaching materials, rather than replacing teachers in any way. An important benefit of using AI-supported question generation with increased efficiency is the potential for personalized approaches towards students.}

\section*{Acknowledgements}
This work was funded by VLAIO (`Flanders Innovation \& Entrepreneurship') in Flanders, Belgium, through the \emph{imec-icon} project AIDA (`AI-Driven e-Assessment'). This research also received funding from the Flemish Government under the “Onderzoeksprogramma Artificiële Intelligentie
(AI) Vlaanderen” programme.
We would like to thank the AIDA partners  Televic Education and  WeZooz Academy for contributing data and use cases.

\bibliography{anthology,custom}

\begin{thebibliography}{35}
\expandafter\ifx\csname natexlab\endcsname\relax\def\natexlab#1{#1}\fi

\bibitem[{Afzal and Mitkov(2014)}]{afzal2014automatic}
Naveed Afzal and Ruslan Mitkov. 2014.
\newblock Automatic generation of multiple choice questions using
  dependency-based semantic relations.
\newblock \emph{Soft Computing}, 18(7):1269--1281.

\bibitem[{Agarwal and Mannem(2011)}]{agarwal2011automatic}
Manish Agarwal and Prashanth Mannem. 2011.
\newblock Automatic gap-fill question generation from text books.
\newblock In \emph{Proceedings of the sixth workshop on innovative use of NLP
  for building educational applications}, pages 56--64.

\bibitem[{Al-Yahya(2011)}]{al2011ontoque}
Maha Al-Yahya. 2011.
\newblock Ontoque: a question generation engine for educational assesment based
  on domain ontologies.
\newblock In \emph{2011 IEEE 11th International Conference on Advanced Learning
  Technologies}, pages 393--395. IEEE.

\bibitem[{Bitew et~al.(2022)Bitew, Hadifar, Sterckx, Deleu, Develder, and
  Demeester}]{bitew2022learning}
Semere~Kiros Bitew, Amir Hadifar, Lucas Sterckx, Johannes Deleu, Chris
  Develder, and Thomas Demeester. 2022.
\newblock Learning to reuse distractors to support multiple choice question
  generation in education.
\newblock \emph{IEEE Transactions on Learning Technologies}.

\bibitem[{Bommasani et~al.(2021)Bommasani, Hudson, Adeli, Altman, Arora, von
  Arx, Bernstein, Bohg, Bosselut, Brunskill
  et~al.}]{bommasani2021opportunities}
Rishi Bommasani, Drew~A Hudson, Ehsan Adeli, Russ Altman, Simran Arora, Sydney
  von Arx, Michael~S Bernstein, Jeannette Bohg, Antoine Bosselut, Emma
  Brunskill, et~al. 2021.
\newblock On the opportunities and risks of foundation models.
\newblock \emph{arXiv preprint arXiv:2108.07258}.

\bibitem[{Brown et~al.(2020)Brown, Mann, Ryder, Subbiah, Kaplan, Dhariwal,
  Neelakantan, Shyam, Sastry, Askell et~al.}]{brown2020language}
Tom Brown, Benjamin Mann, Nick Ryder, Melanie Subbiah, Jared~D Kaplan, Prafulla
  Dhariwal, Arvind Neelakantan, Pranav Shyam, Girish Sastry, Amanda Askell,
  et~al. 2020.
\newblock Language models are few-shot learners.
\newblock \emph{Advances in neural information processing systems},
  33:1877--1901.

\bibitem[{Clark et~al.(2020)Clark, Luong, Le, and Manning}]{Clark2020ELECTRA:}
Kevin Clark, Minh-Thang Luong, Quoc~V. Le, and Christopher~D. Manning. 2020.
\newblock \href {https://openreview.net/forum?id=r1xMH1BtvB} {Electra:
  Pre-training text encoders as discriminators rather than generators}.
\newblock In \emph{International Conference on Learning Representations}.

\bibitem[{Conejo et~al.(2016)Conejo, Guzm{\'a}n, and Trella}]{conejo2016siette}
Ricardo Conejo, Eduardo Guzm{\'a}n, and Monica Trella. 2016.
\newblock The siette automatic assessment environment.
\newblock \emph{International Journal of Artificial Intelligence in Education},
  26(1):270--292.

\bibitem[{Conneau et~al.(2019)Conneau, Khandelwal, Goyal, Chaudhary, Wenzek,
  Guzm{\'a}n, Grave, Ott, Zettlemoyer, and Stoyanov}]{conneau2019unsupervised}
Alexis Conneau, Kartikay Khandelwal, Naman Goyal, Vishrav Chaudhary, Guillaume
  Wenzek, Francisco Guzm{\'a}n, Edouard Grave, Myle Ott, Luke Zettlemoyer, and
  Veselin Stoyanov. 2019.
\newblock Unsupervised cross-lingual representation learning at scale.
\newblock \emph{arXiv preprint arXiv:1911.02116}.

\bibitem[{Daradoumis et~al.(2019)Daradoumis, Puig, Arguedas, and
  Li{\~n}an}]{daradoumis2019analyzing}
Thanasis Daradoumis, Joan Manuel~Marqu{\`e}s Puig, Marta Arguedas, and
  Laura~Calvet Li{\~n}an. 2019.
\newblock Analyzing students' perceptions to improve the design of an automated
  assessment tool in online distributed programming.
\newblock \emph{Computers \& Education}, 128:159--170.

\bibitem[{Davis(2009)}]{davis2009tools}
Barbara~Gross Davis. 2009.
\newblock \emph{Tools for teaching}.
\newblock John Wiley \& Sons.

\bibitem[{Devlin et~al.(2018)Devlin, Chang, Lee, and
  Toutanova}]{devlin2018bert}
Jacob Devlin, Ming-Wei Chang, Kenton Lee, and Kristina Toutanova. 2018.
\newblock Bert: Pre-training of deep bidirectional transformers for language
  understanding.
\newblock \emph{arXiv preprint arXiv:1810.04805}.

\bibitem[{Du et~al.(2017)Du, Shao, and Cardie}]{du-etal-2017-learning}
Xinya Du, Junru Shao, and Claire Cardie. 2017.
\newblock \href {https://doi.org/10.18653/v1/P17-1123} {Learning to ask: Neural
  question generation for reading comprehension}.
\newblock In \emph{Proceedings of the 55th Annual Meeting of the Association
  for Computational Linguistics (Volume 1: Long Papers)}, pages 1342--1352,
  Vancouver, Canada. Association for Computational Linguistics.

\bibitem[{Felice et~al.(2022)Felice, Taslimipoor, and
  Buttery}]{felice-etal-2022-constructing}
Mariano Felice, Shiva Taslimipoor, and Paula Buttery. 2022.
\newblock \href {https://doi.org/10.18653/v1/2022.findings-acl.100}
  {Constructing open cloze tests using generation and discrimination
  capabilities of transformers}.
\newblock In \emph{Findings of the Association for Computational Linguistics:
  ACL 2022}, pages 1263--1273, Dublin, Ireland. Association for Computational
  Linguistics.

\bibitem[{Goto et~al.(2010)Goto, Kojiri, Watanabe, Iwata, and
  Yamada}]{goto2010automatic}
Takuya Goto, Tomoko Kojiri, Toyohide Watanabe, Tomoharu Iwata, and Takeshi
  Yamada. 2010.
\newblock Automatic generation system of multiple-choice cloze questions and
  its evaluation.
\newblock \emph{Knowledge Management \& E-Learning: An International Journal},
  2(3):210--224.

\bibitem[{Hill and Simha(2016)}]{hill2016automatic}
Jennifer Hill and Rahul Simha. 2016.
\newblock Automatic generation of context-based fill-in-the-blank exercises
  using co-occurrence likelihoods and google n-grams.
\newblock In \emph{Proceedings of the 11th Workshop on Innovative Use of NLP
  for Building Educational Applications}, pages 23--30.

\bibitem[{Lee and Seneff(2007)}]{lee2007automatic}
John Lee and Stephanie Seneff. 2007.
\newblock Automatic generation of cloze items for prepositions.
\newblock In \emph{Eighth Annual Conference of the International Speech
  Communication Association}.

\bibitem[{Malinova and Rahneva(2016)}]{malinova2016automatic}
Anna Malinova and Olga Rahneva. 2016.
\newblock Automatic generation of english language test questions using
  mathematica.
\newblock In \emph{CBU International Conference Proceedings}, volume~4, pages
  906--909.

\bibitem[{Marrese-Taylor et~al.(2018)Marrese-Taylor, Nakajima, Matsuo, and
  Yuichi}]{marrese2018learning}
Edison Marrese-Taylor, Ai~Nakajima, Yutaka Matsuo, and Ono Yuichi. 2018.
\newblock Learning to automatically generate fill-in-the-blank quizzes.
\newblock \emph{arXiv preprint arXiv:1806.04524}.

\bibitem[{Mitkov et~al.(2006)Mitkov, Le~An, and Karamanis}]{mitkov2006computer}
Ruslan Mitkov, Ha~Le~An, and Nikiforos Karamanis. 2006.
\newblock A computer-aided environment for generating multiple-choice test
  items.
\newblock \emph{Natural language engineering}, 12(2):177--194.

\bibitem[{Oller~Jr(1973)}]{oller1973cloze}
John~W Oller~Jr. 1973.
\newblock Cloze tests of second language proficiency and what they measure 1.
\newblock \emph{Language learning}, 23(1):105--118.

\bibitem[{Ouyang et~al.(2022)Ouyang, Wu, Jiang, Almeida, Wainwright, Mishkin,
  Zhang, Agarwal, Slama, Ray et~al.}]{ouyang2022training}
Long Ouyang, Jeffrey Wu, Xu~Jiang, Diogo Almeida, Carroll Wainwright, Pamela
  Mishkin, Chong Zhang, Sandhini Agarwal, Katarina Slama, Alex Ray, et~al.
  2022.
\newblock Training language models to follow instructions with human feedback.
\newblock \emph{Advances in Neural Information Processing Systems},
  35:27730--27744.

\bibitem[{Papasalouros et~al.(2008)Papasalouros, Kanaris, and
  Kotis}]{papasalouros2008automatic}
Andreas Papasalouros, Konstantinos Kanaris, and Konstantinos Kotis. 2008.
\newblock Automatic generation of multiple choice questions from domain
  ontologies.
\newblock \emph{e-Learning}, 1:427--434.

\bibitem[{Perez-Beltrachini et~al.(2012)Perez-Beltrachini, Gardent, and
  Kruszewski}]{perez2012generating}
Laura Perez-Beltrachini, Claire Gardent, and German Kruszewski. 2012.
\newblock Generating grammar exercises.
\newblock In \emph{The 7th Workshop on Innovative Use of NLP for Building
  Educational Applications, NAACL-HLT Worskhop 2012}, pages 147--157.

\bibitem[{Pino et~al.(2008)Pino, Heilman, and Eskenazi}]{pino2008selection}
Juan Pino, Michael Heilman, and Maxine Eskenazi. 2008.
\newblock A selection strategy to improve cloze question quality.
\newblock In \emph{Proceedings of the Workshop on Intelligent Tutoring Systems
  for Ill-Defined Domains. 9th International Conference on Intelligent Tutoring
  Systems, Montreal, Canada}, pages 22--32.

\bibitem[{Pugh et~al.(2016)Pugh, De~Champlain, Gierl, Lai, and
  Touchie}]{pugh2016using}
Debra Pugh, Andre De~Champlain, Mark Gierl, Hollis Lai, and Claire Touchie.
  2016.
\newblock Using cognitive models to develop quality multiple-choice questions.
\newblock \emph{Medical teacher}, 38(8):838--843.

\bibitem[{Raffel et~al.(2020)Raffel, Shazeer, Roberts, Lee, Narang, Matena,
  Zhou, Li, and Liu}]{raffel2020exploring}
Colin Raffel, Noam Shazeer, Adam Roberts, Katherine Lee, Sharan Narang, Michael
  Matena, Yanqi Zhou, Wei Li, and Peter~J Liu. 2020.
\newblock Exploring the limits of transfer learning with a unified text-to-text
  transformer.
\newblock \emph{The Journal of Machine Learning Research}, 21(1):5485--5551.

\bibitem[{Slavuj et~al.(2021)Slavuj, Prskalo, and
  Bakaric}]{slavuj2021automatic}
Vanja Slavuj, L~Nacinovic Prskalo, and M~Brkic Bakaric. 2021.
\newblock Automatic generation of language exercises based on a universal
  methodology: An analysis of possibilities.
\newblock \emph{Bulletin of the Transilvania University of Brasov. Series IV:
  Philology and Cultural Studies}, pages 29--48.

\bibitem[{Stasaski and Hearst(2017)}]{stasaski2017multiple}
Katherine Stasaski and Marti~A Hearst. 2017.
\newblock Multiple choice question generation utilizing an ontology.
\newblock In \emph{Proceedings of the 12th Workshop on Innovative Use of NLP
  for Building Educational Applications}, pages 303--312.

\bibitem[{Sumita et~al.(2005)Sumita, Sugaya, and
  Yamamoto}]{sumita2005measuring}
Eiichiro Sumita, Fumiaki Sugaya, and Seiichi Yamamoto. 2005.
\newblock Measuring non-native speakers’ proficiency of english by using a
  test with automatically-generated fill-in-the-blank questions.
\newblock In \emph{Proceedings of the second workshop on Building Educational
  Applications Using NLP}, pages 61--68.

\bibitem[{Sun et~al.(2018)Sun, Zhu, Xiao, Xiao, and Wei}]{sun2018automatic}
Bo~Sun, Yunzong Zhu, Yongkang Xiao, Rong Xiao, and Yungang Wei. 2018.
\newblock Automatic question tagging with deep neural networks.
\newblock \emph{IEEE Transactions on Learning Technologies}, 12(1):29--43.

\bibitem[{Susanti et~al.(2017)Susanti, Tokunaga, Nishikawa, and
  Obari}]{susanti2017evaluation}
Yuni Susanti, Takenobu Tokunaga, Hitoshi Nishikawa, and Hiroyuki Obari. 2017.
\newblock Evaluation of automatically generated english vocabulary questions.
\newblock \emph{Research and practice in technology enhanced learning},
  12(1):1--21.

\bibitem[{Taylor(1953)}]{taylor1953cloze}
Wilson~L Taylor. 1953.
\newblock “cloze procedure”: A new tool for measuring readability.
\newblock \emph{Journalism quarterly}, 30(4):415--433.

\bibitem[{Wang et~al.(2020)Wang, Yao, Kwok, and Ni}]{wang2020generalizing}
Yaqing Wang, Quanming Yao, James~T Kwok, and Lionel~M Ni. 2020.
\newblock Generalizing from a few examples: A survey on few-shot learning.
\newblock \emph{ACM computing surveys (csur)}, 53(3):1--34.

\bibitem[{Wei et~al.(2022)Wei, Wang, Schuurmans, Bosma, brian ichter, Xia, Chi,
  Le, and Zhou}]{wei2022chain}
Jason Wei, Xuezhi Wang, Dale Schuurmans, Maarten Bosma, brian ichter, Fei Xia,
  Ed~H. Chi, Quoc~V Le, and Denny Zhou. 2022.
\newblock \href {https://openreview.net/forum?id=_VjQlMeSB_J} {Chain of thought
  prompting elicits reasoning in large language models}.
\newblock In \emph{Advances in Neural Information Processing Systems}.

\end{thebibliography}
\bibliographystyle{acl_natbib}
\begin{figure*}[h!]
\includegraphics[width=\textwidth]{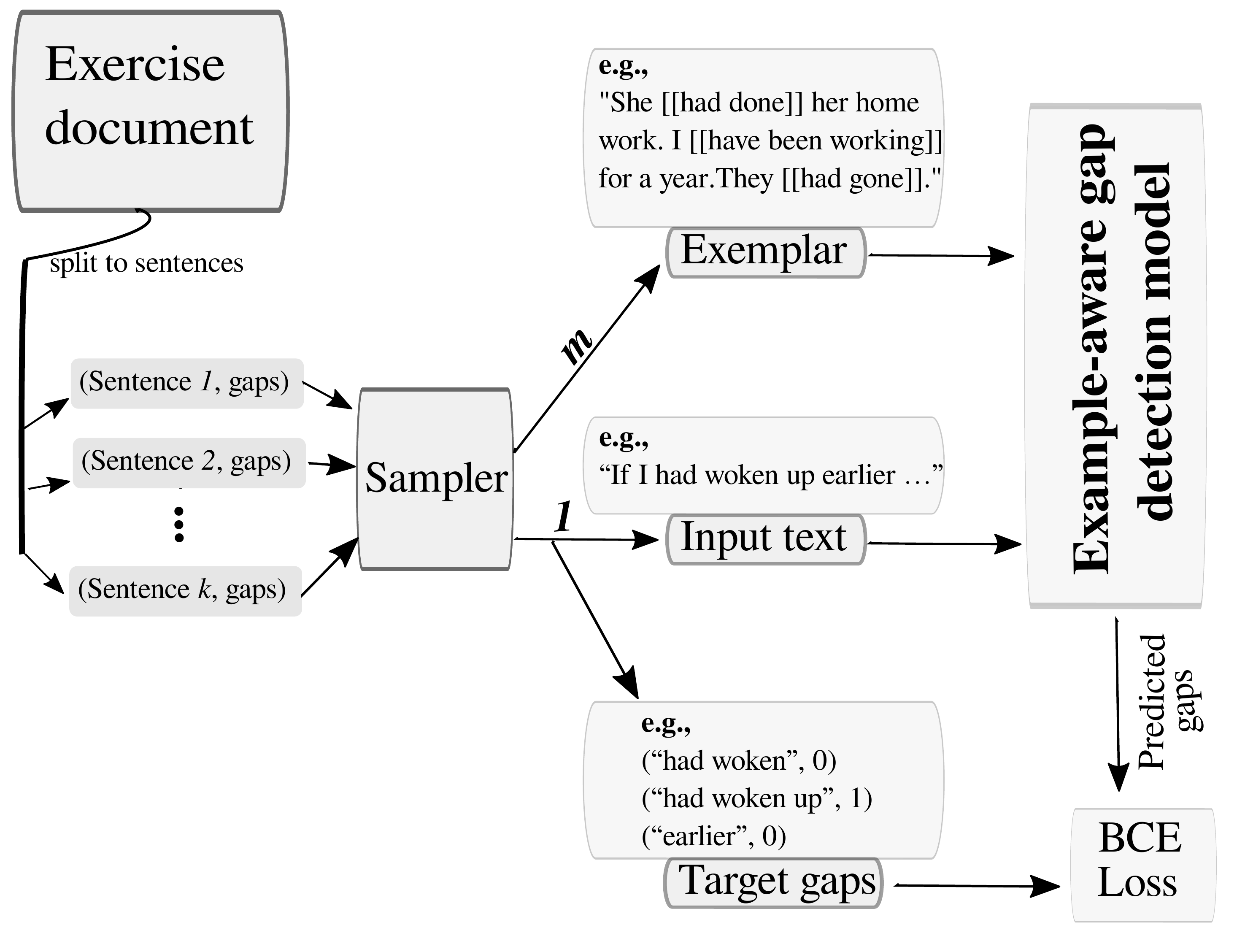}
\caption{Training procedure of our example-aware gap detection model. First, we split exercise documents into list of sentences. Then we create (input, exemplar) training pairs that will be used by our model. We use one sentence as an input, while the exemplar is made up of sentences that are uniformly sampled from the remaining sentences. The exemplar is constructed by concatenating the $m$ sampled sentences. The special symbols ``[['' and ``]]'' in the exemplar indicate the gap positions. Binary cross entropy (BCE) loss is used to train our models.  } \label{fig:training-details}
\end{figure*}
\appendix

\section{Training details}
\label{sec:appendix}
In this section we detail our training procedure. As depicted in \figref{fig:training-details}, we first split training exercises into list of sentences, along with their corresponding gap position indications. In order to create new (input, exemplar) pair, we sample $1$ sentence from the sentence list to be used as our \emph{input} text, and we uniformly sample $1$ up to $m$ (we set $m=3$) sentences from the remaining sentence list to be used as our exemplar. We form our exemplar by concatenating all the sampled sentences with gap positions indicated by special tokens ``[['' and ``]]''. Then our model is trained by minimizing the binary cross entropy (BCE) loss between predicted gaps and their target labels (1 for valid gaps, and 0 otherwise). 

\end{document}